\documentclass[10pt,twocolumn,letterpaper]{article}

\usepackage{cvpr}              

\usepackage{multirow} 
\usepackage{colortbl}  
\usepackage{xcolor}
\usepackage{array}
\usepackage{lipsum}
\usepackage{arydshln}
\usepackage{tikz}
\usepackage{etoolbox}
\makeatletter
\robustify\@latex@warning@no@line
\makeatother
\usepackage{authblk}
\usepackage{fancyhdr}

\newcommand{\circled}[2][]{\tikz[baseline=(char.base)]
    {\node[shape = circle, draw, inner sep = 1pt]
    (char) {\phantom{\ifblank{#1}{#2}{#1}}};%
    \node at (char.center) {\makebox[0pt][c]{#2}};}}
\robustify{\circled}
\definecolor{lightlightgray}{rgb}{0.9, 0.9, 0.9}

\definecolor{cvprblue}{rgb}{0.21,0.49,0.74}
\usepackage[pagebackref,breaklinks,colorlinks,allcolors=cvprblue]{hyperref}

\def\confName{CVPR}
\def\confYear{2025}

\title{FADA: Fast Diffusion Avatar Synthesis with Mixed-Supervised Multi-CFG Distillation}

\author{
\textbf{Tianyun Zhong$^1$\thanks{Equal Contribution}~\thanks{Done during an internship at ByteDance.}~, Chao Liang$^{2*}$~, Jianwen Jiang$^{2*}$~ \thanks{Project Lead}~, Gaojie Lin$^2$~, Jiaqi Yang$^2$~, Zhou Zhao$^1$}
} 
\affil{
$^1$Zhejiang University, $^2$Bytedance Intelligent Creation\\ 
\footnotesize \texttt{zhongtianyun@zju.edu.cn}
\\
{\footnotesize \texttt{\{liangchao.0412,jianwen.alan,yjq850207131\}@gmail.com}}
} 
\affil{
\url{https://fadavatar.github.io/}
}

\begin{document}
\maketitle
\begin{abstract}

Diffusion-based audio-driven talking avatar methods have recently gained attention for their high-fidelity, vivid, and expressive results. However, their slow inference speed limits practical applications. Despite the development of various distillation techniques for diffusion models, we found that naive diffusion distillation methods do not yield satisfactory results. Distilled models exhibit reduced robustness with open-set input images and a decreased correlation between audio and video compared to teacher models, undermining the advantages of diffusion models.
To address this, we propose FADA (Fast Diffusion Avatar Synthesis with Mixed-Supervised Multi-CFG Distillation). We first designed a mixed-supervised loss to leverage data of varying quality and enhance the overall model capability as well as robustness. Additionally, we propose a multi-CFG distillation with learnable tokens to utilize the correlation between audio and reference image conditions, reducing the threefold inference runs caused by multi-CFG with acceptable quality degradation. Extensive experiments across multiple datasets show that FADA generates vivid videos comparable to recent diffusion model-based methods while achieving an NFE speedup of \textbf{4.17-12.5} times. Demos are available at our webpage \href{https://fadavatar.github.io}{https://fadavatar.github.io}.
\end{abstract}

\section{Introduction}
\label{sec:intro}

Talking avatar synthesis aims to animate a given portrait image using driven video or audio. Due to the ease of obtaining audio inputs and their low usage barrier, there has been a surge of recent work in this area.

Based on the technical route, talking avatar synthesis methods can be categorized into GAN-based\cite{zhou2020makelttalk, wang2021one, zhou2021pose, zhao2022thin, hong2022depth, ma2023styletalk, guo2024liveportrait}, NeRF-based\cite{guo2021ad,tang2022real,ye2023geneface,ye2023geneface++,ye2024real3d,ye2024mimictalk}, and the more recent diffusion model-based approaches\cite{tian2024emo,chen2024echomimic,xu2024hallo,cui2024hallo2,jiang2024loopy,lin2024cyberhost,wang2024v}. Diffusion model-based methods, in particular, have gained attention due to their strong generative capabilities. These models exhibit excellent robustness in input images and produce vivid and highly expressive videos that naturally align with the audio.

However, the current diffusion model-based talking head generation methods suffer from slow inference speeds, which hinder their practical application. Methods\cite{jiang2024loopy,xu2024hallo,chen2024echomimic} based on Stable Diffusion (SD)\cite{rombach2022high}, all require multiple denoising steps. Moreover, to ensure the correlation between the audio input and the image input in the final video, multi-CFG (Classifier-Free Guidance) inference is often employed, further increasing the runtime. Although most methods achieve 30-40 denoising steps with the help of DDIM\cite{song2020denoising}, they still face speed challenges due to CFG calculations.

Recently, several works have explored diffusion distillation in text-to-image \cite{sauer2023adversarial, li2024snapfusion, nguyen2024swiftbrush, dao2025swiftbrush, lin2024sdxl, luo2023latent, song2023consistency, zheng2024trajectory} and text-to-video tasks \cite{lv2024fastercache} to accelerate inference. Nonetheless, there has been no direct exploration of diffusion distillation for talking avatar synthesis tasks. Our observations indicate that simply applying text-to-image and text-to-video diffusion distillation methods results in a noticeable decline in performance.

Unlike text to image or video tasks, talking avatar synthesis involves two key conditions: audio and a reference image. Since audio has a relatively weaker influence on video generation\cite{jiang2024loopy,tian2024emo}, the final output is jointly influenced by both conditions, necessitating careful parameter tuning in multi-CFG settings. The sensitivity to condition guidance often leads to severe artifacts when applying diffusion distillation methods with few-step inference. 

In this paper, we propose \textbf{FADA} (\textbf{FA}st \textbf{D}iffusion \textbf{A}vatar Synthesis with Mixed-Supervised Multi-CFG Distillation) to address these issues from two perspectives, resulting in an audio-driven talking avatar diffusion model that balances both speed and quality.

First, we address the robustness of the distilled model. In this task, the mapping between audio and portrait movement must be learned from data. Previous works have emphasized meticulous data selection \cite{jiang2024loopy, lin2024cyberhost, chen2024echomimic}, such as removing instances with excessive movement or asynchronous audio, to ensure high audio-portrait movement correlation. This stringent selection process excludes a significant amount of data that could enhance robustness. Considering that a well-trained teacher model already provides high audio-portrait correlation guidance, this discarded data can be utilized in training the student model. We designed a mixed-supervised learning strategy that adaptively adjusts the weights between learning from the teacher model's guidance and the student model's own learning from previously discarded data.
Secondly, we address the speed issues of multimodal conditions in the distillation process. In audio-driven methods, the final output is achieved through composite CFG inference \cite{jiang2024loopy, lin2024cyberhost, xu2024hallo}. To make this process transparent to the student model during training, we introduce a set of learnable tokens. These tokens mimic the weighted calculations of multi-CFG with its coefficients and are injected into the network as control signals, enabling the student model to better learn the relationships between multiple conditions, including those present during inference. Since the learnable token conditions replicate the multi-CFG inference calculation process, we can reduce the number of multi-CFG model inferences during actual inference, further decreasing inference time on top of the denoising step distillation.

Based on the above considerations and designed methods, we implemented FADA, a diffusion distillation method for audio-driven talking head generation, achieving up to a 12.5$\times$ speed-up while generating vivid portrait videos comparable to recent undistilled diffusion models. In summary, the contributions of this paper include:

\begin{enumerate} 
\item We propose \textbf{FADA}, the first diffusion-based distillation framework for audio-driven talking avatar tasks. It includes a mixed-supervised loss to learn from data of varying quality and learnable token conditions to mimic the multi-CFG inference process, reducing performance loss by distillation and multi-CFG inferences.
\item Quantitative and qualitative experiments demonstrate that FADA matches state-of-the-art generation quality while achieving significantly higher efficiency. \end{enumerate}
\section{Related Works}
\label{sec:related_works}

\subsection{Audio-Driven Talking Avatar}
\label{sec:related_works_audio_driven}
The task of audio-driven talking avatars has gained increasing attention recently, propelled by advancements in video-generation technologies. GAN-based methods \cite{zhou2020makelttalk, wang2021one, zhou2021pose, zhao2022thin, hong2022depth, ma2023styletalk, guo2024liveportrait} typically employ a two-stage pipeline involving an audio-to-motion module and a motion-to-video module. Sadtalker \cite{zhang2023sadtalker} introduces a 3DMM-based motion representation along with a conditional VAE for generating high-fidelity talking heads. LivePortrait \cite{guo2024liveportrait} expands the training dataset incorporating images and videos, using implicit keypoints for high-quality generation. NeRF-based methods \cite{guo2021ad, tang2022real, ye2023geneface, ye2023geneface++, ye2024real3d, ye2024mimictalk} explore animating talking heads in 3D neural spaces. Real3D-Portrait \cite{ye2024real3d} combines a pre-trained image-to-triplane module with a motion adapter for zero-shot and one-reference 3D talking head generation. MimicTalk \cite{ye2024mimictalk} captures the talking style of individuals through in-context-learning audio-to-motion modules and rapid fine-tuning of triplanes for mimic talking head generation.

Diffusion-based talking avatar synthesis has swiftly become a crucial and efficient approach \cite{tian2024emo,chen2024echomimic,xu2024hallo,cui2024hallo2,jiang2024loopy,lin2024cyberhost,wang2024v}. EMO \cite{tian2024emo} pioneers a double UNet framework in audio-driven avatar creation, yielding impressive and lifelike results. Concurrently, EchoMimic \cite{chen2024echomimic} trains on audio and facial landmarks for versatile control inputs. Hallo \cite{xu2024hallo} incorporates a hierarchical audio-driven visual synthesis module to refine audio-visual alignments, while Hallo2 \cite{cui2024hallo2} explores image augmentation strategies for 4K avatar video synthesis. Loopy \cite{jiang2024loopy} introduces a motion-frame squeeze technique and audio-to-latent training for expressive results. Despite their progress in expressive audio-driven generation, diffusion-based methods still face challenges with slow inference speeds.

\subsection{Diffusion Model Acceleration}
\label{sec:related_diff_dist}

Diffusion acceleration involves lowering costs for denoising runs and reducing the number of runs. Various denoising methods aim to cut costs, such as utilizing Model quantization ~\cite{shang2023post,so2024temporal,he2024ptqd,li2024q,sui2024bitsfusion,zhao2024vidit} to decrease the model parameters directly. ViDiT-Q [47] introduced a new diffusion transformer quantization framework for video and image with mixed-precision quantization. DeepCache [20] takes cues from video compression to propagate reusable result regions through the model's internal structure, while Faster Diffusion [12] emphasizes re-using computed encoder features.

Reducing the number of denoising runs is a direct way to accelerate diffusion models. SD-Turbo\cite{sauer2023adversarial} leverages the ideas from GAN and introduces adversarial diffusion distillation to enhance the generation fidelity. SnapFusion\cite{li2024snapfusion} proposes progressive distillation to achieve fewer steps by learning the previous student model progressively. SwiftBrush\cite{nguyen2024swiftbrush} and SwiftBrushV2\cite{dao2025swiftbrush} utilizing variational score distillation to make it possible for texts-only distillation in text-to-image tasks. SDXL-lightning\cite{lin2024sdxl} mixes progressive distillation and adversarial distillation to achieve one-step image generation with a delicate training process. Applying consistency models~\cite{luo2023latent,song2023consistency} to text-to-image tasks also helps reduce diffusion inference steps. LCM\cite{luo2023latent} views the guided reverse diffusion process as solving an augmented probability flow ODE. TCD\cite{zheng2024trajectory} proposes trajectory consistency distillation to accurately trace the entire trajectory of the probability flow ODE in semi-linear form with an exponential integrator. However, there is no specially-designed distillation framework and techniques for talking avatar synthesis tasks to our knowledge.

\section{Methodology}
\label{sec:methodology}
\begin{figure*}[t]
    \centering
    \includegraphics[width=\textwidth]{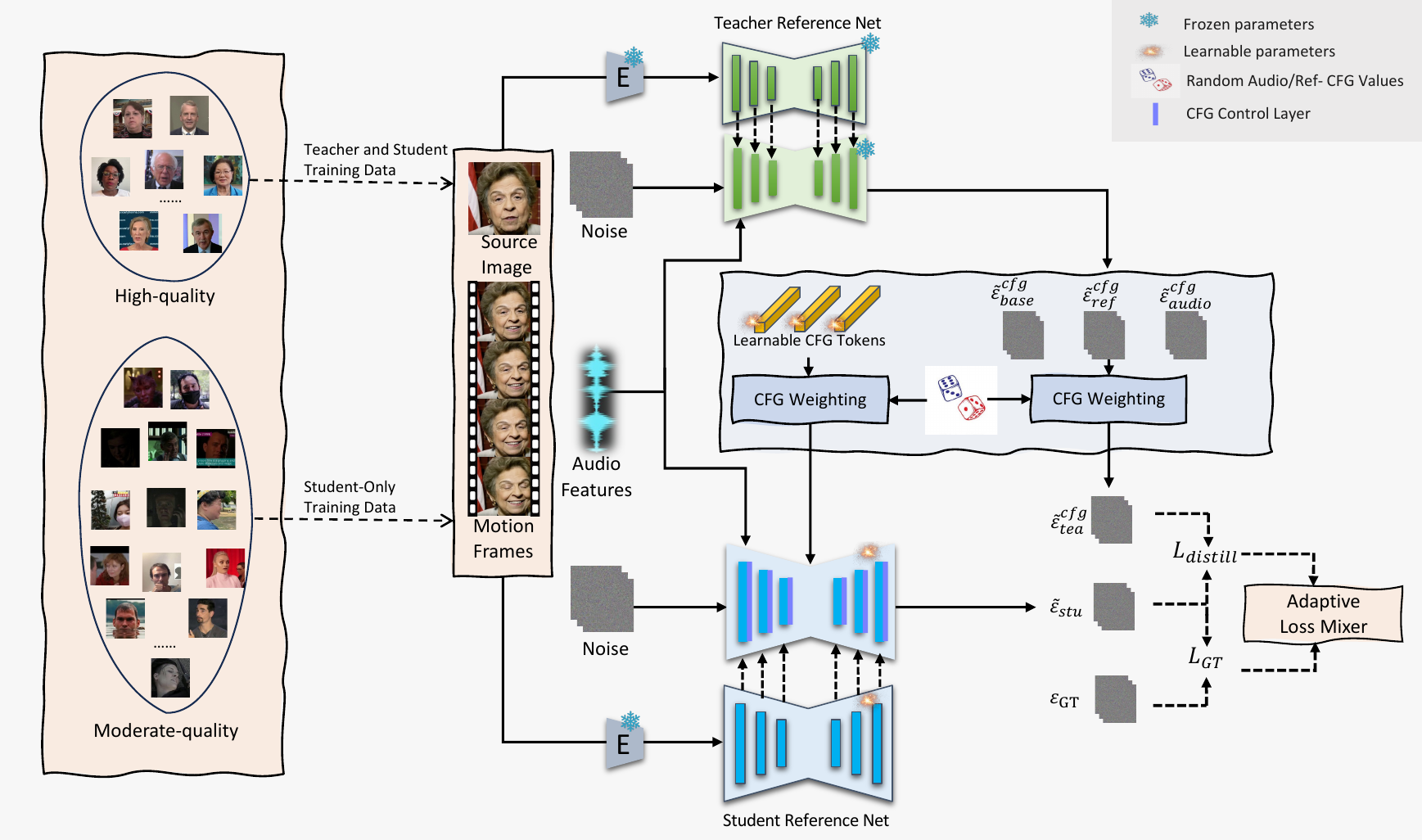}
    \caption{\textbf{Overall distillation framework of FADA.} The teacher model is trained only with high-quality data, which is omitted in the figure. The student model is trained by a mixed loss of ground-truth and teacher-supervised loss to leverage data of varying quality. Learnable token-based CFG conditions enable the student model to mimic the multi-CFG process, further reducing inference times. For simplicity, we have omitted some components commonly used in previous methods.
}
    \label{fig:framework}
    \vspace{-0.2cm}
\end{figure*}

In this section, we introduce our proposed method, FADA, a diffusion distillation framework for avatar synthesis, as illustrated in Figure \ref{fig:framework}. Section \ref{sec:overall_framework} provides an overview of the diffusion distillation framework, based on recent PeRFlow \cite{yan2024perflow}. Section \ref{sec:adaptive} delves into the specifics of the adaptive mixed-supervised distillation process. In Section \ref{sec:multiple-cfg}, we present the multiple-CFG distillation technique with a learnable token implementation. 

\subsection{Overall Framework and Preliminary}
\label{sec:overall_framework}
Building upon cutting-edge diffusion-based methods for talking avatar synthesis \cite{hu2024animate, tian2024emo, xu2024hallo, cui2024hallo2, wang2024v, chen2024echomimic, jiang2024loopy, lin2024cyberhost}, we introduce a dual-Unet architecture for the teacher model, leveraging pre-trained weights from Stable Diffusion\cite{rombach2022high} 1.5. As depicted in Figure~\ref{fig:framework}, the reference image and motion frames (i.e., images from the last clip) undergo processing by a fixed VAE \cite{kingma2013auto} and reference net. The output from each reference layer is then input to the corresponding layer in the denoising net for spatial attention. To facilitate audio-driven controls and ensure temporal consistency in talking avatar synthesis, audio attention and temporal attention layers are integrated into the denoising net. Initially following the Loopy design \cite{jiang2024loopy}, we simplify the model by removing the TSM and Audio2Latents modules, streamlining subsequent model distillation. The student model mirrors the teacher's structure, with the addition of a CFG control layer, detailed in Section \ref{sec:multiple-cfg}.

We choose the recent PeRFlow \cite{yan2024perflow} as the optimization target for distillation. Initially, we train the teacher model on carefully curated high-quality datasets that prioritize motion amplitude and audio-visual synchronization. The teacher model is optimized using the $\epsilon$-prediction DDPM loss \cite{ho2020denoising}, where it predicts noise $\epsilon_{\mathrm{tea}}(\cdot)$ from noisy latent $z_{t}$ at timestep $t$. The training objective can be succinctly described as:
\begin{equation}
    \label{eq:teacher_objective}
    L_{\mathrm{tea}} = \mathbb{E}_{z_t, c, t, \epsilon}\left[\lVert \epsilon - \epsilon_{\mathrm{tea}}(z_t, t, c;\theta_{\mathrm{tea}}) \rVert_{2}^{2} \right],
\end{equation}

\noindent where $c$ includes the audio feature, reference image and motion frames. 

Note that it is not necessary for the teacher model to be trained with flow-matching loss, and both $\epsilon$-prediction and $\upsilon$-prediction are valid for PeRFlow distillation. During the student distillation training, firstly same-interval $K$ time windows $\{[t_k,t_{k-1})\}_{k=K}^{1}$ are created where $t_k=k/K,k\in\{0,1,...,K\}$. A time window $[t_k,t_{k-1})$ is randomly sampled out, and the standpoint $z_{t_{k}}$ can be derived from the marginal distribution of the ground-truth targets, where $z_{t_{k}}=\sqrt{1-\sigma^2(t_k)} z_0 + \sigma(t_k) \epsilon$ and $\sigma(t)$ denotes the noise schedule. Then the teacher-predicted ending point $\hat{z}_{t_{k-1}}$ of this time window can be predicted and solved by ODE solver $\Phi(\cdot)$:
\begin{equation}
    \label{eq:perflow_end}
    \hat{z}_{t_{k-1}} = \Phi(z_{t_{k}}, t_k, t_{k-1};\theta_{\mathrm{tea}}),
\end{equation}

\noindent where $\Phi(\cdot)$ denotes the DDIM solver\cite{song2020denoising} which performs several inference steps iteratively to reach the endpoint of this time window. Note that the parameters of the teacher model are frozen and no gradient computations are needed here. After that, the input noisy latent of student $\hat{z}_t$ will be derived from a linear interpolation between starting point $t_k$ and ending point $t_{k-1}$:
\vspace{-0.1cm}
\begin{equation}
    \label{eq:perflow_interpolation}
    \hat{z}_{t} = \frac{z_{t_{k}}-\hat{z}_{t_{k-1}}}{t_{k}-t_{k-1}}(t-t_k).
\end{equation}
\vspace{-0.3cm}

Through the linear interpolation above, the ODE flow will be rectified to piecewise straight lines so only a few steps are enough to denoising in a time window, which reduces the number of timesteps at inference stage. Meanwhile, we can figure out the target noise $\hat{\epsilon}(t)$ via parameterization\cite{yan2024perflow}:
\vspace{-0.1cm}
\begin{equation}
    \label{eq:perflow_parameterization}
    \hat{\epsilon}(t) = \frac{\hat{z}_{t_{k-1}} - \lambda_{k} z_{t_{k}}}{\eta_{k}},
\end{equation}
\vspace{-0.1cm}
\noindent Where $\lambda_k=\sqrt{\alpha_{t_{k-1}}}/\sqrt{\alpha_{t_{k}}}$ and $\eta_k=\sqrt{1-\alpha_{t_{k-1}}} - \sqrt{1-\alpha_{t_{k}}} \sqrt{\alpha_{t_{k-1}}}/\sqrt{\alpha_{t_{k}}}$ respectively. Finally, the training objective of the distillation process can be obtained as
\begin{equation}
    \label{eq:student_objective}
    L_{\mathrm{distill}} = \mathbb{E}_{k,t\in[t_k,t_{k-1}),z_t, c,\epsilon}\left[\lVert \hat{\epsilon}(t)-\epsilon_{\mathrm{stu}}(\hat{z_t}, t, c;\theta_{\mathrm{stu})} \rVert_{2}^{2} \right],
\end{equation}
\noindent where $\epsilon_{\mathrm{stu}}(\cdot)$ refers to the predicted noise from the student model and $L_{\mathrm{distill}}$ represents the distillation loss. Note that $L_{\mathrm{distill}}$ is only a trivial part of our framework, the mixed-supervised training and multi-CFG distillation will be proposed in Section \ref{sec:adaptive} and \ref{sec:multiple-cfg}.

\subsection{Adaptive Mixed-Supervised Distillation}
\label{sec:adaptive}

In our experiments, we found that distillation significantly degrades the model's performance, especially overall video synthesis quality, as shown in Table \ref{tab:aba_mixed}. Increasing data can improve performance, but in audio-driven portrait generation, the weak control of audio over motion requires the model to learn high-quality motion patterns to fill in details not covered by audio. This necessitates stricter data filtering, reducing the available data quantity.

Fortunately, this problem can be mitigated in our proposed distillation process. In our design, we first maintain the teacher training unchanged with only a high-quality dataset $\mathcal{A}$ to ensure its output stability. Then during the distillation stage, the data filtering strategies will be relaxed and some moderate-quality data samples ( a larger dataset $\mathcal{B}$ that includes $\mathcal{A}$ where $ |\mathcal{B}|>|\mathcal{A}|$.) will be fed to student training.  Since the student model mimics the teacher model's output, it can maintain the high-quality generation capability of the teacher model even with the inclusion of lower-quality data. To further leverage the new data during the distillation stage, we introduce a ground-truth supervised loss $L_{\mathrm{gt}}$. The ground-truth noisy latent $z^{*}_{t}$ and ground-truth target noise $\epsilon^{*}(t)$ can be simply obtained by replacing $\hat{z}_{t_{k-1}}$ with ground truth ending point $z_{t_{k-1}}$ in Formula \ref{eq:perflow_interpolation} and \ref{eq:perflow_parameterization}. Similar to distillation loss computation, the ground-truth loss $L_{\mathrm{GT}}$ and total mixed loss $L_{\mathrm{total}}$ can be described as
\begin{equation}
    \label{eq:gt_objective}
    L_{\mathrm{GT}} = \mathbb{E}_{k,t\in[t_k,t_{k-1}),z_t, c,\epsilon}\left[\lVert \epsilon^{*}(t)-\epsilon_{\mathrm{stu}}(z_t^{*}, t, c;\theta_{\mathrm{stu})} \rVert_{2}^{2} \right],
\end{equation}

\vspace{-0.2cm}
\begin{equation}
    \label{eq:total_objective}
    L_{\mathrm{total}} = L_{\mathrm{distill}} + \mathcal{W} \cdot L_{\mathrm{GT}},
\end{equation}

\noindent where $\mathcal{W}$ represents the loss weight of $L_{GT}$. Through this approach, the student model can better learn high-quality motion pattern generation from the teacher model while also learning more common patterns from a broader range of data to enhance its generalization capability. A trivial implementation idea is to set a fixed hyper-parameter to $\mathcal{W}$ but this is not very effective because it assumes every sample has the same learning value, which is inaccurate.

Hence, an adaptive method to balance teacher-supervised loss $L_{\mathrm{teacher}}$ and ground-truth-supervised loss $L_{\mathrm{gt}}$ is necessary, and the potential information behind the two losses should be taken into account. Considering the ratio $\mathcal{R}=L_{\mathrm{gt}}/L_{\mathrm{teacher}}$, it indicates the difference level between the $L_{\mathrm{gt}}$ and $L_{\mathrm{teacher}}$. Based on qualitative observations of the $\mathcal{R}$ value, there are two trends: as $\mathcal{R}$ increases, the sample is more likely to be a valuable learning case, such as a singing video, since it is often outside the teacher-training dataset $\mathcal{A}$. However, if $\mathcal{R}$ is too large, the sample is more likely to be low-quality, such as one with poor audio-visual sync, because the teacher results are well synced. 

Therefore, we should gradually increase the loss weight when $\mathcal{R}$ increases in a reasonable range, and also punish the loss weight if $\mathcal{R}$ increases out of the peak threshold $\mathcal{R}_{p}$. For those cases whose $\mathcal{R}$ is bigger than the dead threshold $\mathcal{R}_{d}$, the ground-truth supervised loss weight should be zero. The detailed formulation for loss weight $W$ can be obtained as: 
\vspace{-0.4cm}

\begin{equation}
\label{eq:loss_weight}
\mathcal{W}=\mathcal{W}_0\cdot \left\{
             \begin{array}{lr}
             \mathcal{R}^s, 
             &  \mathcal{R} \in [0, \mathcal{R}_{p})
             \\
             \mathcal{R}_{p}^{s} (\mathcal{R}_{d}-\mathcal{R})/(\mathcal{R}_{d}-\mathcal{R}_{p}),
             &\mathcal{R}\in [\mathcal{R}_{p}, \mathcal{R}_{d})
             \\
             0, &  \mathcal{R} \in [\mathcal{R}_{d},+\infty)
             \end{array}
\right.
\end{equation}
\noindent where $s$, $\mathcal{W}_0$, $\mathcal{R}_{p}$ and $\mathcal{R}_{d}$ are hyper-parameters, they are set to 0.25, 0.2, 30, 100 respectively in this paper. Note that these hyper-parameters are relatively robust across different data distributions because they are used to select data within a relatively moderate range for training.

Through the proposed design, FADA learns general generation capabilities from more ordinary data during distillation, enhancing model robustness. Our experiments validate the proposed loss's effectiveness. Additionally, our adaptive mixed-supervised distillation can extend to tasks like text-to-image or text-to-video without specific restrictions.

\subsection{Multi-CFG Distillation with Learnable Token}
\label{sec:multiple-cfg}

In audio-driven talking avatar synthesis tasks, the Multi-CFG technique is utilized \cite{jiang2024loopy, lin2024cyberhost} to achieve robust control of multiple conditions (reference image and audio) effectively, 
Specifically, the single CFG calculates the direction from unconditional predicted values to a certain conditional predicted value and then moves a certain distance from the unconditional predicted values to obtain conditional predicted values. Furthermore, the Multi-CFG recursively computes the next CFG based on the first CFG. In this paper, multi-CFG inference result noise $\epsilon_{cfg}$ can be derived as follows\cite{jiang2024loopy}:
\begin{equation}
    \label{eq:naive_multi_cfg}
    \hat{\epsilon}_{cfg} = \mathrm{cfg}_{a} \times (\hat{\epsilon}_{a} - \hat{\epsilon}_{r}) + \mathrm{cfg}_{r} \times (\hat{\epsilon}_{r} - \hat{\epsilon}_{b}) + \hat{\epsilon}_{b},
\end{equation}
\noindent where $\mathrm{cfg}_{a}$ and $\mathrm{cfg}_{r}$ indicate the CFG guidance scale of audio and reference condition. $\hat{\epsilon}_{a}$ refers to the noise prediction with both audio and reference conditions, while $\hat{\epsilon}_{r}$ removes the audio condition and $\hat{\epsilon}_{b}$ lacks both audio and reference. 

Therefore, during inference, the model needs to run three times for each denoising step, which is time-consuming.

It naturally comes to mind to inject the CFG guidance scale as a condition into the network, and then let the student model learn the CFG reasoning characteristics of the teacher model during the distillation process, which echoes a similar concept to the $\omega$-condition \cite{meng2023distillation}. Specifically, the teacher model will conduct complete CFG reasoning during training to obtain predictions $\hat{\epsilon}_{\mathrm{cfg}}$ controlled by CFG. Subsequently, these predictions are further integrated into Formulas \ref{eq:perflow_end}, \ref{eq:perflow_interpolation} and \ref{eq:perflow_parameterization} to derive input noisy latent $\hat{z}_t^{\mathrm{cfg}}$ and target noise $\hat{\epsilon}^{\mathrm{cfg}}(t)$ controlled by CFG, eventually undergoing a similar distillation procedure as Formula \ref{eq:student_objective}.

However, directly adding the CFG value as a condition to the network does not yield satisfactory results. A careful design is needed for the injection method of the CFG scale condition. By observing the calculation of Multi-CFG reasoning in Formula \ref{eq:naive_multi_cfg}, it is evident that this is akin to performing linear transformations on three independent vectors $\epsilon_{b}$, $\epsilon_{a}$, and $\epsilon_{r}$ in the noise space using the given CFG strengths $\mathrm{cfg}_{a}$ and $\mathrm{cfg}_{r}$. Meanwhile, $\epsilon_{r}$ serves as a crucial link between the $\epsilon_{b}$ and $\epsilon_{a}$. Simply embedding the CFG values overlooks the aforementioned relationships. Therefore, we introduce learnable tokens $\gamma_{b}$, $\gamma_{r}$, and $\gamma_{r}$ into the model to learn the features required for CFG strength control and help the student model mimic the multi-CFG process. To achieve this, we design the CFG embedding $Emb_{\mathrm{cfg}}$ of the CFG guidance scale $\mathrm{cfg}_{a}$ and $\mathrm{cfg}_{r}$ as follows:
\vspace{-0.4cm}

\begin{equation}
    \label{eq:embedding_multi_cfg}
    Emb_{\mathrm{cfg}} = \mathrm{cfg}_{a} \times (\gamma_{a} - \gamma_{r}) + \mathrm{cfg}_{r} \times (\gamma_{r} - \gamma_{b}) + \gamma_{b}.
\end{equation}

After obtaining the CFG embedding $Emb_{\mathrm{cfg}}$, we similarly introduce a CFG layer after each audio layer in the denoising network, injecting the information from $Emb_{\mathrm{cfg}}$ into the denoising process through cross attention. Importantly, the CFG layer does not introduce excessive performance overhead, especially when compared to the original cost of running three CFG runs. Therefore, our proposed multi-CFG distillation with learnable tokens can achieve nearly three times faster operation speed.

\section{Experiments}
\label{sec:experiments}
In this section, we will introduce the experimental results. In Section \ref{sec:experimental_settings}, we will present the details of the implementation of FADA and the meta-information of our training and testing datasets. In Section \ref{sec:comparison_with_sota}, we will compare our proposed methods with other state-of-the-art diffusion-based talking avatar methods via quantitative and qualitative comparison. In Section \ref{sec:ablation}, we will analyze our proposed techniques including the basic distillation method, adaptive mixed-supervised distillation and multi-CFG distillation, and prove their effectiveness.

\subsection{Experimental Settings}
\label{sec:experimental_settings}

\paragraph{Datasets}

Regarding the training dataset, we primarily used speaker videos obtained from the internet, sliced and cropped to durations ranging from two to ten seconds with facial frames, and resized to 512$\times$512 resolution. We designed multi-dimensional filters for this dataset, including filters for data source, image quality, audio-visual synchronization, background stability and degree of movement, each with different thresholds to control the filtering intensity. As elucidated in Section \ref{sec:adaptive}, by using strict filtering thresholds, we obtained approximately 160 hours of high-quality training data $\mathcal{A}$, which was utilized in teacher pre-training. Additionally, by using lenient filtering thresholds, we obtained around 1300 hours of moderate-quality training data $\mathcal{B}$ for distillation training. Following the approach of Echomimic \cite{chen2024echomimic}, we selected 100 random samples from the open-source test sets HDTF \cite{zhang2021flow} and CelebV-HQ \cite{zhu2022celebv} to validate the model's performance on various real human scenarios. Furthermore, to evaluate the model's performance in more open and extreme scenarios, we adopted the openset test set from Loopy \cite{jiang2024loopy} for qualitative comparisons, which contains various kinds of portraits and audios.

\vspace{-0.6cm}
\paragraph{Implementation Details} 

The training process consists of teacher pre-training and student distillation. The training process of the teacher model closely aligns with the first two stages of recent methods\cite{tian2024emo,jiang2024loopy} with high-quality dataset $\mathcal{A}$. In student distillation with the moderate-quality dataset $\mathcal{B}$, the student model is initialized from the teacher model and integrated with CFG learnable tokens and CFG layers. During testing, the CFG guidance scale is set to 2.5 and 6.5 for reference and audio conditions in models without CFG distillation, while the scales were set to 2.0 and 6.5 for models with CFG distillation. 

The number of time windows in PeRFlow\cite{yan2024perflow} is 4. All results of FADA are conducted with 6-step inferences, where the two time-windows near $T=1$ contain two inference steps while the others contain one inference step.

\subsection{Comparison with State-of-the-Arts}
\label{sec:comparison_with_sota}

\begin{table*}[t]

  \setlength{\tabcolsep}{3pt}
  \centering
  \caption{\small Quantitative comparisons with state-of-the-art methods on CelebV-HQ \& HDTF test sets. \textbf{Bold} and \underline{underlined} numbers refer to the best and the second-best result in this test set.}
  \label{tab:com_cele_hdtf}
  
    \begin{tabular}{ccccccccccccc}
    \toprule
    \multirow{2}{*}{\textbf{Method}}  &  \multirow{2}{*}{\textbf{NFE-D}} &  \multicolumn{5}{c}{\textbf{CelebV-HQ Test}} & \multicolumn{5}{c}{\textbf{HDTF Test}} \\ \cmidrule(lr){3-7} \cmidrule(lr){8-12}
                                 & & \textbf{IQA}$\uparrow$ & \textbf{Sync-D}$\downarrow$ &  
                                 \textbf{FVD-R}$\downarrow$ & \textbf{FID}$\downarrow$ & \textbf{E-FID}$\downarrow$  & \textbf{IQA}$\uparrow$  & \textbf{Sync-D}$\downarrow$ &  
                                 \textbf{FVD-R}$\downarrow$ & \textbf{FID}$\downarrow$ & \textbf{E-FID}$\downarrow$ 
    \\ \midrule
    Sadtalker\cite{zhang2023sadtalker}
    & - & $2.953^{\pm.086}$ & 8.765 & 171.8 & 36.64 & 2.248
    & $3.435^{\pm.069}$ & 7.870 & 24.93 & 25.35 & 1.559 \\
    Hallo\cite{xu2024hallo} 
    & 80 & $3.505^{\pm.091}$ & 9.079 & 53.99 & 35.96 & 2.426 
    & $3.922^{\pm.079}$ & 7.917 & 21.71 & 20.15 & 1.337 \\
    EchoMimic\cite{chen2024echomimic} 
    & 60 & $3.307^{\pm.104}$ & 10.37 & 54.71 & 35.37 & 3.018 
    & $3.994^{\pm.083}$ & 9.391 & 18.71 & 19.01 & 1.328 \\
    V-Express\cite{wang2024v} 
    & 50 & $2.946^{\pm.123}$ & 9.415 & 117.8 & 65.09 & 2.414
    & $3.482^{\pm.071}$ & 7.382 & 48.03 & 30.91 & 1.506\\
    \rowcolor{lightlightgray} Ours-Balanced 
    & \underline{18} & \textbf{$3.588^{\pm.091}$}  & 8.998 & \textbf{50.36} & \textbf{33.67} & \textbf{2.202}
    & \underline{$3.951^{\pm.082}$} & 7.925 & \textbf{13.32} & \textbf{18.39} & 1.362\\

    \rowcolor{lightlightgray} Ours-Fast 
    & \textbf{6} & \underline{$3.550^{\pm.093}$} & \textbf{8.746} & 54.69 & \underline{35.01} & 2.604
    & $3.927^{\pm.081}$ & 7.996 & \underline{16.67} & \underline{18.51} & 1.635\\

    \bottomrule
    \end{tabular}%
\end{table*}

\paragraph{Metrics and Baselines}
For the purpose of quantitative comparisons, we evaluate the generated talking avatar videos with several metrics. We assess the image quality with IQA\cite{wu2023q} metrics and evaluate the audio-visual synchronization distance with a pre-trained SyncNet\cite{prajwal2020lip} as Sync-D. FID and FVD\cite{unterthiner2019fvd} (with 3D ResNet) are assessed to evaluate the image/video-level fidelity difference between generated results and ground truths in the image inception space. E-FID (Expression-FID)\cite{tian2024emo} measures the inception distance between the generated and ground truths in expression parameter space. NFE-D indicates the number of function evaluations of the denoising net. For instance, our teacher model conducts 25-step DDIM inference with 3 CFG runs, leading to 75 NFE-D.

Since our proposed FADA is a diffusion distillation framework, we mainly compare recent open-source diffusion-based state-of-the-art methods in talking portrait generation, including Hallo\cite{xu2024hallo},  EchoMimic\cite{chen2024echomimic}, and V-Express\cite{wang2024v}. Additionally, for reference, we also include the state-of-the-art GAN-based method SadTalker\cite{zhang2023sadtalker}.

\vspace{-0.4cm}
\paragraph{Quantitative Comparison}
As shown in Table \ref{tab:com_cele_hdtf}, we conduct quantitative comparison experiments with CelebV-HQ and HDTF test sets. Since the differences in IQA are subtle, we provide 95$\%$ confidence intervals to show their statistical significance. \textbf{Ours-Balanced} refers to our proposed method with mixed-supervised distillation with PeRFlow loss. Based on it, \textbf{Ours-Fast} utilizes multi-CFG distillation to achieve about 3-time speedup and gets a slight performance drop. Ours-Balanced achieves better IQA, FVD, FID, and E-FID on the challenging open-set CelebVHQ test set compared to the baselines. It also leads or is comparable in most metrics on the simpler HDTF test set. Ours-Fast maintains similar performance in IQA and Sync-D metrics with a slight drop in FVD, FID, and E-FID, which is still comparable to the baselines in most metrics. In terms of audio synchronization, the trend is similar. Notably, VExpress results show almost only lip movements, which, despite good metrics, do not produce satisfactory video quality. Finally, in terms of speedup, it is evident that both Ours-Balanced and Ours-Fast achieve significant speedup using the same SD base model as the baselines.

\vspace{-0.4cm}
\paragraph{Qualitative Comparison}
\begin{figure*}[t]
    \centering
    \includegraphics[width=0.95\textwidth]{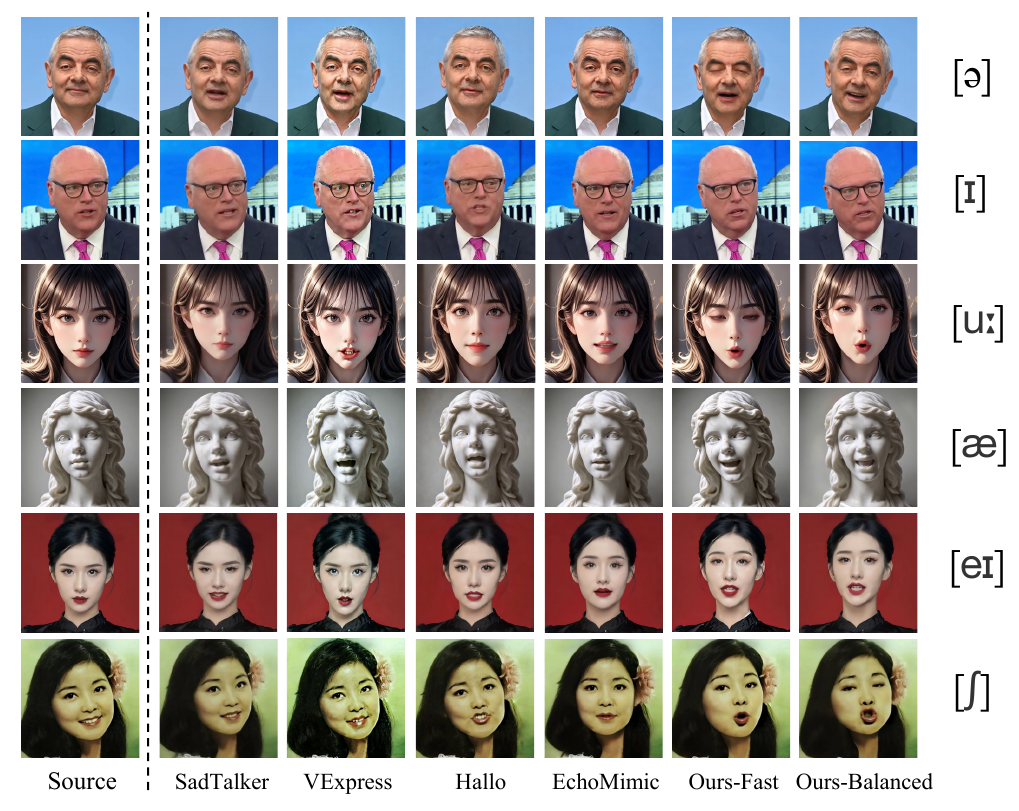}
    \vspace{-0.4cm}
    \caption{\textbf{Qualitative comparisons between FADA and baselines across different portraits and pronunciations in openset.} 
}
    \label{fig:vis_comp}
    \vspace{-0.4cm}
\end{figure*}
As illustrated in Figure \ref{fig:vis_comp} and in the supplementary video materials, Ours-Fast and Ours-Balanced maintain good video quality and vivid audio-visual expressiveness. We highly recommend readers watch the supplementary videos since it is difficult for the static image to show the audio-motion synchronization and temporal consistency. V-Express shows color differences and inferior identity-preserving abilities in all cases. Sadtalker generates good identity-preserving videos at the image level, but the head motions and lip movements are not well-synced with the emotion and pace of the given audio. EchoMimic and Hallo generate acceptable expressions, but there are temporal jitters that damage the visual quality. Moreover, by observing case-4 and case-6, we can find that FADA produces more full and plump pronunciations than baselines.

\subsection{Ablation Study}
\label{sec:ablation}

\begin{table}[t]

  \small
  \footnotesize
  \setlength{\tabcolsep}{2pt}
  \centering
  \caption{\small Ablation study of basic distillation methods on HDTF test set. All methods in this table are free from mixed-supervised and CFG distillation.}
  \label{tab:aba_basic}
  
    \begin{tabular}{ccccccc}
    \toprule

    \textbf{Method} & \textbf{NFE-D} & \textbf{IQA}$\uparrow$ & \textbf{Sync-D}$\downarrow$  &  
                                 \textbf{FVD-R}$\downarrow$ & \textbf{FID}$\downarrow$ & \textbf{E-FID}$\downarrow$ \\
    \midrule
    Teacher with $\mathcal{A}$ & 75 & \textbf{3.998} & \textbf{7.849} & \textbf{18.27} & \textbf{18.49} & \textbf{1.365} \\
    Teacher with $\mathcal{B}$ & 75 & \underline{3.885} & 8.123 & \underline{19.64} & 21.40 & 1.689 \\
    \hdashline
    TCD & 18 & 3.827 & \underline{7.937} & 23.01 & 23.15 & 1.893 \\
    RFlow & 18 & 2.836 & 8.066 & 81.10 & 49.73 & 2.484 \\
    \rowcolor{lightlightgray} PeRFlow  & 18 & 3.823 & 8.001 & 21.91 & \underline{21.26} & \underline{1.477} \\  
    
    \bottomrule
    \end{tabular}%
\end{table}

\begin{table}[t]

  \small
  \footnotesize
  \setlength{\tabcolsep}{1pt}
  \centering
  \caption{\small Ablation study of adaptive mixed-supervised distillation on HDTF test set. All methods use PeRFlow with dataset $\mathcal{B}$ without CFG distillation. Fixed-x indicates fixed ground-truth loss with weight x. Adaptive-unlimited-s and Adaptive-s indicate adaptive ground-truth loss with hyper-parameter $s$, with the former being without peak and dead threshold.}
  \vspace{-0.2cm}
  \label{tab:aba_mixed}
  
    \begin{tabular}{ccccccc}
    \toprule

    \textbf{Method} & \textbf{NFE-D} & \textbf{IQA}$\uparrow$ & \textbf{Sync-D}$\downarrow$  &  
                                 \textbf{FVD-R}$\downarrow$ & \textbf{FID}$\downarrow$ & \textbf{E-FID}$\downarrow$ \\
    \midrule
    Teacher with $\mathcal{A}$ & 75 & \textbf{3.998} & \textbf{7.849} & 18.27 & \underline{18.49} & \underline{1.365} \\
    Stu. with $\mathcal{A}$ & 18 & 3.823 & 8.001 & 21.91 & 21.26 & 1.477 \\
    
    Stu. with $\mathcal{B}$ (Fixed-0) & 18 & 3.762 & 7.909 & 25.06 & 21.47 & 1.462\\
    \hdashline
    + Fixed-0.2 & 18 & 3.863 & 7.934 & \underline{13.71} & 19.94 & 1.434\\
    + Fixed-1.0 & 18 & 3.812 & \underline{7.903} & 24.53 & 21.42 & 1.494\\
    + Fixed-0.2, only $\mathcal{A}$ & 18 & 3.880 & 7.962 & 15.32 & 20.71 & 1.512\\
    + Adaptive-unl.-0.5 & 18 & 3.824 & 7.914 & 15.15 & 20.74 & 1.471\\
    + Adaptive-unl.-0.25 & 18 & 3.846 & 7.906 & 14.20 & 19.92 & 1.424\\
    + Adaptive-0.5 & 18 & 3.949 & 7.919 & 13.83 & 18.64 & 1.395\\
    \rowcolor{lightlightgray} + Adaptive-0.25 & 18 & \underline{3.951} & 7.925 & \textbf{13.32} & \textbf{18.39} & \textbf{1.362}\\
    
    \bottomrule
    \end{tabular}%
\end{table}

\begin{table}[t]

  \small
  \footnotesize
  \setlength{\tabcolsep}{1pt}
  \centering
  \caption{\small Ablation study of multi-CFG distillation on HDTF test set. All methods in this table are PeRFlow + mixed-supervised distillation. CFG time embed refers to CFG distillation through timestep embedding injection.}
  \vspace{-0.2cm}
  \label{tab:aba_cfg}
  
    \begin{tabular}{ccccccc}
    \toprule

    \textbf{Method} & \textbf{NFE-D} & \textbf{IQA}$\uparrow$ & \textbf{Sync-D}$\downarrow$  &  
                                 \textbf{FVD-R}$\downarrow$ & \textbf{FID}$\downarrow$ & \textbf{E-FID}$\downarrow$ \\
    \midrule
    Teacher with $\mathcal{A}$ & 75 & \textbf{3.998} & \textbf{7.849} & \underline{18.27} & \textbf{18.49} & \textbf{1.365} \\
    \hdashline
    w.o. CFG  & 6 & 3.674 & 9.020 & 27.10 & 26.56 & 2.096 \\
    + CFG time embed & 6 & 3.701 & 8.577 & 23.24 & 22.47 & 1.791 \\
    + CFG layer  & 6 & 3.746 & \underline{7.906} & 20.05 & 20.94 & \underline{1.620}\\
    \rowcolor{lightlightgray}+ CFG layer w. tokens  & 6 & \underline{3.927} & 7.996 & \textbf{16.67} & \underline{18.51} & 1.635\\

    \bottomrule
    \end{tabular}%
    \vspace{-0.2cm} 
\end{table}

\paragraph{Analysis of Distillation Methods.}

As illustrated in Table \ref{tab:aba_basic}, we first compared various basic methods of diffusion distillation, including TCD \cite{zheng2024trajectory}, RFlow \cite{liu2022rectified}, and PeRFlow \cite{yan2024perflow}. It was observed that compared to the TCD method, the PeRFlow method showed similar performance in IQA, FVD, Sync-D, and FID metrics, but with a significant advantage in E-FID. Additionally, the basic RFlow method could not handle the 6-step inference in our task setting. Therefore, we selected PeRFlow as the basic distillation method.
We also tried simply adding moderate-quality data during teacher model training, which led to a decrease in model performance. These results indicate that existing distillation methods cannot directly produce a high-quality student model. Additionally, attempting to enhance the teacher model's capabilities by relaxing data filtering criteria and increasing data volume does not yield positive results.

\vspace{-0.4cm}
\paragraph{Analysis of Adaptive Mixed-Supervised Distillation}

Here, based on the previously selected PeRFlow, we added dataset $\mathcal{B}$ and a mixed-supervised loss to investigate the effectiveness of our Adaptive Mixed-Supervised Distillation. 
As shown in Table \ref{tab:aba_cfg}, below the hdashline, we provide the performance of Mixed-Supervised Distillation with the addition of dataset $\mathcal{B}$ under different parameters. Comparing the effects of different Fixed-x values, it is evident that the ground-truth loss improves model performance, and a smaller loss weight value of 0.2 yields better results. 
We further investigated the effectiveness of the adaptive loss weight strategy. In the adaptive-unl. setting (unlimited, without using peak and dead thresholds), the adaptive loss weight increases infinitely as $\mathcal{R}$ increases, resulting in slightly inferior performance compared to the Fixed-0.2 setting. However, when employing the complete adaptive strategy, the ground-truth loss of low-quality samples is automatically discarded, leading to a distillation effect that surpasses the Fixed setting and approaches the performance of the teacher model, especially in terms of the FVD metrics. We also compared different values of $s$ and ultimately chose 0.25 due to its slight advantage.

Additionally, by comparing the performance of teacher and student models under different data qualities and scales, it can be demonstrated that mixed supervised distillation actually enhances the model's utilization of moderate-quality data. As depicted in Tables \ref{tab:aba_basic} and \ref{tab:aba_mixed}, when the teacher model is trained with a large-scale moderate-quality dataset, a decline in model performance is observed, whereas the student model exhibits the opposite trend. Briefly, only mixed-supervised distillation can take advantage of the moderate-quality data.

\paragraph{Analysis of Multi-CFG Distillation}

\begin{figure}[t]
    
    \centering
    \vspace{-0.2cm}
    \includegraphics[width=0.45\textwidth]{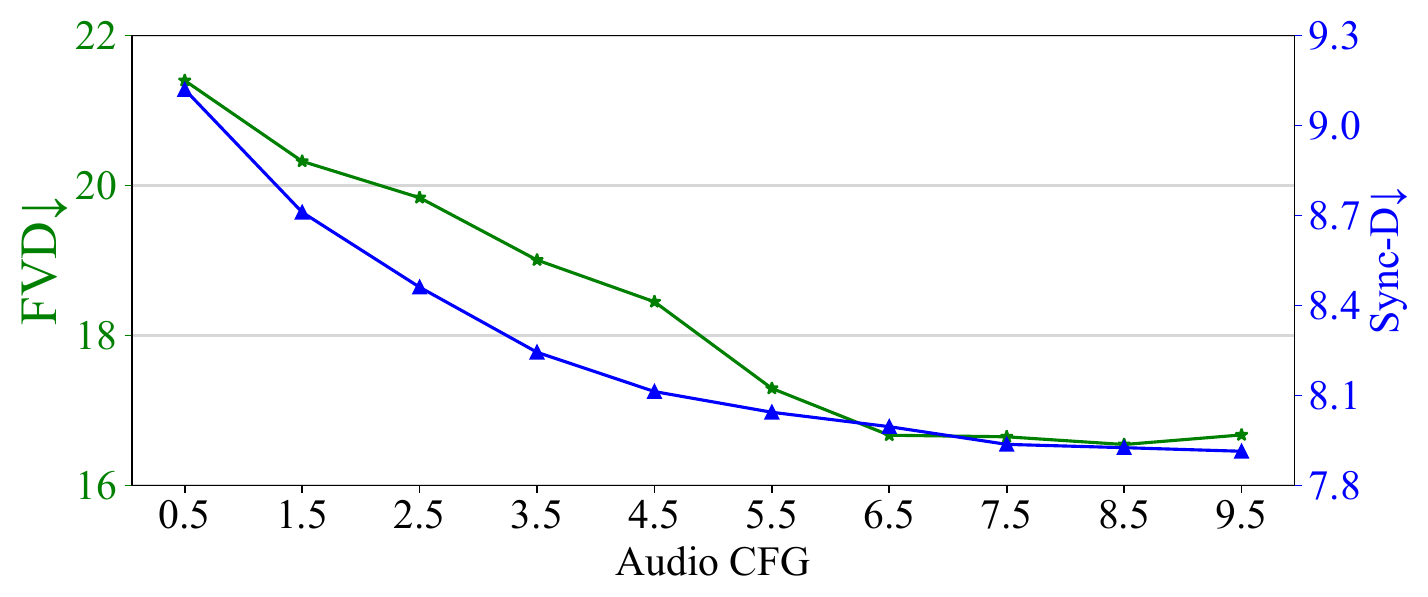}
    \vspace{-0.2cm}
    \caption{\textbf{Line charts showing the variations of FVD and Sync-D metrics with different audio CFG on HDTF test set.} Reference CFG is set to 2.0 in this figure. 
}
    \label{fig:audiocfg}
    \vspace{-0.4cm}
\end{figure}

\begin{figure}[t]
    \centering
    \includegraphics[width=0.45\textwidth]{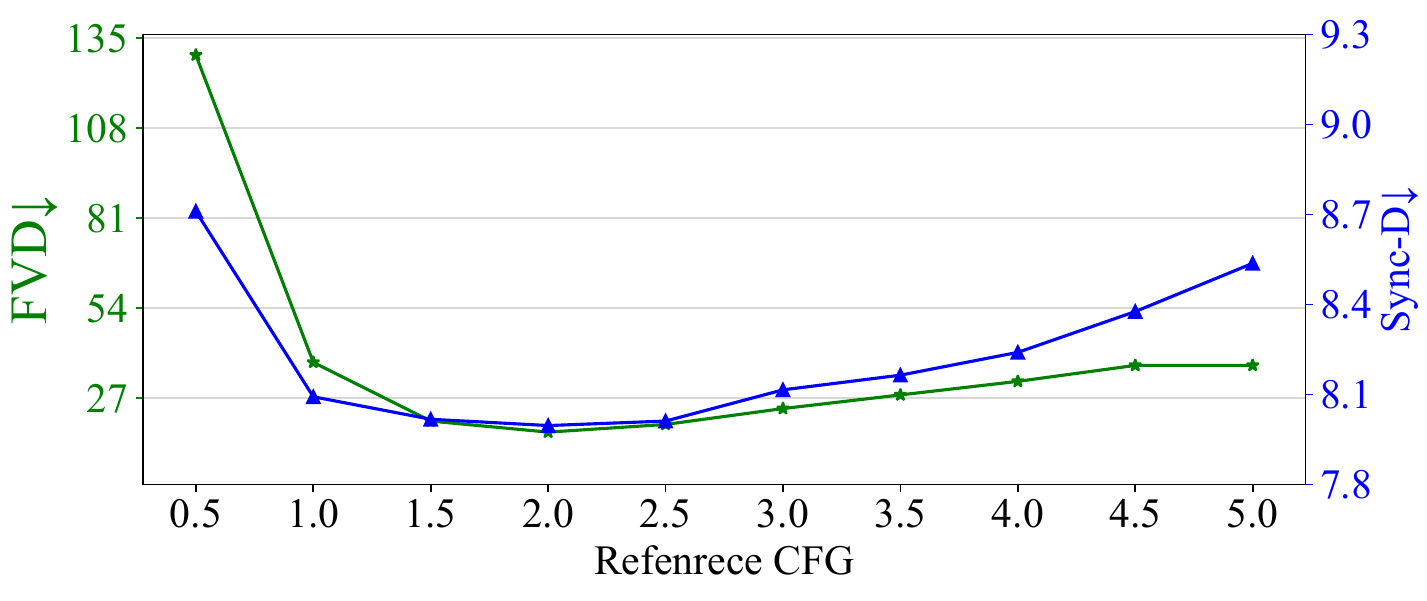}
    \vspace{-0.2cm}
    \caption{\textbf{Line charts showing the variations of FVD and Sync-D metrics with different ref CFG on HDTF test set.} Audio CFG is set to 6.5 in this figure.
}
    \label{fig:refcfg}
    \vspace{-0.4cm}
\end{figure}

\vspace{-0.4cm}
As demonstrated in Table \ref{tab:aba_cfg}, we further explore the effects of multi-CFG distillation on top of mixed-supervised distillation. Initially, we tested the results of conditional reasoning without CFG distillation, which significantly deteriorated the model performance, especially leading to a doubling of the FVD metric representing video quality. For techniques like time embedding injection similar to $\omega$-condition \cite{meng2023distillation}, they partially restored model performance but with limited efficacy. In contrast, our proposed learnable token-based CFG control layers significantly enhance model performance, nearly matching the teacher model and even surpassing it in FVD. Notably, our model requires only 8\% of the teacher model's NFE. We also tested without learnable tokens, and while the pure CFG control layer had some effect, it could not match the teacher model, especially in overall video quality metrics like FVD.

To verify whether our proposed multi-CFG distillation effectively captures the influence of audio and reference conditions on the generated results, we plotted line charts of FVD and Sync-D metrics with variations in audio CFG and reference CFG, as shown in Figures \ref{fig:audiocfg} and \ref{fig:refcfg}. 
First, we observed improvements in both FVD and Sync-D as audio CFG increased, reaching a stable maximum value after 6.5. Hence, we ultimately chose 6.5 as the audio CFG for inference. 
Second, the influence of reference CFG on FVD and Sync-D metrics displayed an initially positive, then negative trend, with optimal performance achieved around the 2.0 position. 
These results indicate that our CFG control layer effectively mimics the multi-CFG process and achieves good results.
\section{Conclusion}
\label{sec:conclusion}
In summary, we proposed FADA, a fast diffusion avatar synthesis framework with mixed-supervised multi-CFG distillation. By designing a mixed-supervised loss, we leverage data of varying quality to enhance the robustness of generated results. Furthermore, a learnable token-based multi-CFG condition design is introduced to maintain the correlation between the audio and the generated video in the distilled models. The quantitative and qualitative experiments across multiple datasets show that our balanced setting achieves fast inference and high-fidelity generation ability. Ablation studies prove our proposed methods are effective in FADA. The limitations, future works and ethical concerns of this paper will be discussed in Appendix B, H and I respectively.

{
    \small
    \bibliographystyle{ieeenat_fullname}
    \bibliography{main}
}



\end{document}